\documentclass{article}

\usepackage[preprint]{corl_2026} 

\usepackage{amsmath,amssymb}
\usepackage{booktabs}
\usepackage{multirow}
\usepackage{graphicx}
\usepackage{xcolor}
\usepackage{subcaption}
\usepackage{pdfpages}
\usepackage{float}
\usepackage{placeins}
\usepackage{tikz}
\usetikzlibrary{arrows.meta,positioning,fit,calc,shapes.geometric}

\newcommand{\ours}{X-NavDP}

\title{\textbf{X-NavDP: Generalizing Navigation Diffusion Policy to Novel Behavior and Embodiments with Group Q-score Reweighted Matching}}

\author{
  \textbf{Tianyu Yang}$^{1,2\ast}$\quad
  \textbf{Yiming Zeng}$^{3,2\ast}$\quad
  \textbf{Wenzhe Cai}$^{2\ast}$\quad
  \textbf{Yuqiang Yang}$^2$\quad
  \textbf{Jiaqi Peng}$^{4,2}$\\[0.2em]
  \textbf{Hui Cheng}$^3$\quad
  \textbf{Jiangmiao Pang}$^2$\quad
  \textbf{Tai Wang}$^{2\dagger}$\\[0.85em]
  $^1$Fudan University \quad
  $^2$Shanghai AI Laboratory \quad
  $^3$Sun Yat-sen University\\[0.28em]
  $^4$Tsinghua University\\[0.2em]
  $^\ast$Equal contribution. \quad
  $^\dagger$Corresponding author.
}

\begin{document}
\maketitle

\vspace{-0.3cm}
\noindent\begin{minipage}{\linewidth}
    \centering
    \includegraphics[width=\linewidth]{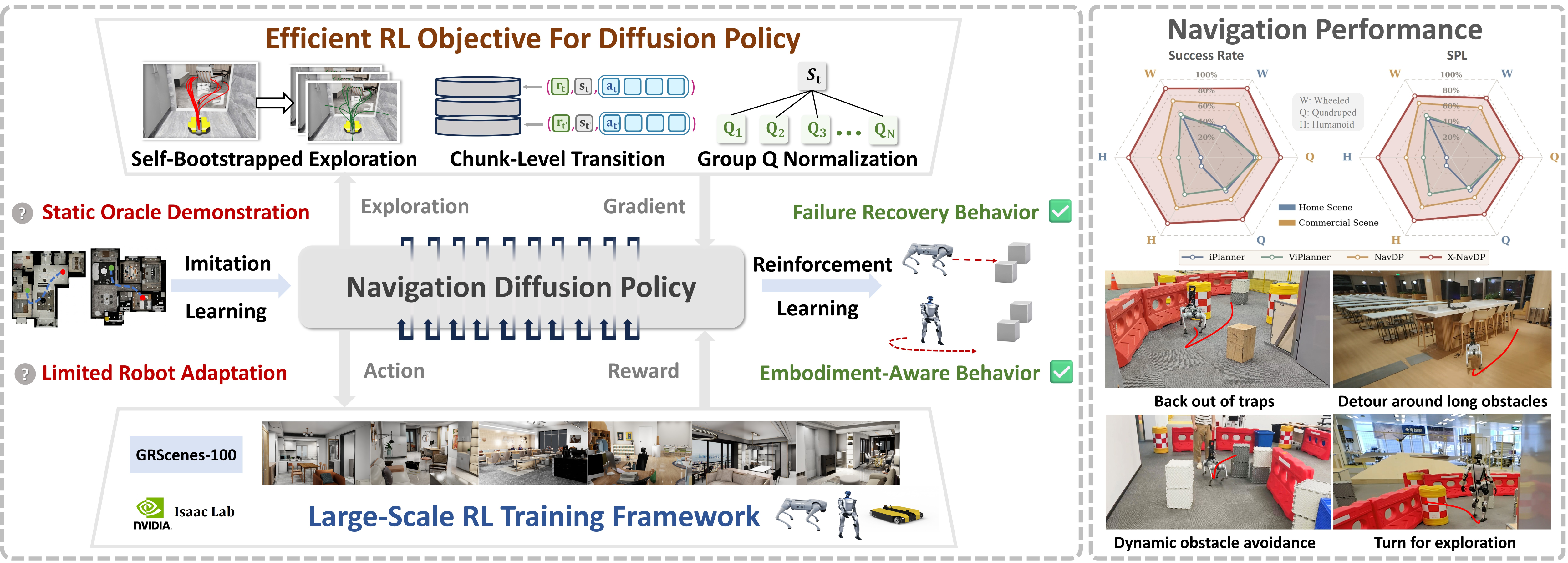}
    \captionof{figure}{We develop an RL post-training framework that improves pretrained navigation diffusion policies in general navigation and obstacle avoidance, strengthens cross-embodiment navigation, and enables new skills such as failure recovery.}
    \label{fig:teaser}
\end{minipage}
\vspace{-0.35cm}

\begin{abstract}
    Pretraining navigation diffusion policies rely on large-scale expert demonstrations. 
    These data are typically generated by a fully-informed oracle planner suited to a single nominal robot. 
    This limits the policy's generalization to diverse embodiments and challenging scenarios (e.g., escaping dead ends or detouring long obstacles) that demand diverse local reactive behaviors with only onboard local observations.
    Post-training the policy with reinforcement learning (RL) offers a principled remedy.
    However, previous RL for diffusion approaches lead to only marginal improvements.
    This is because the intractable likelihood of diffusion policies renders policy gradients unstable in addition to inefficient policy exploration.
    To address these challenges, we propose a data-efficient diffusion RL post-training framework - GQRM (Group Q-score Reweighted Matching). Our framework introduces two complementary designs: (i) a self-bootstrapped exploration strategy with behavior perturbation that preserves the pretrained policy prior, and (ii) a group Q-score normalization mechanism that computes per-trajectory values on each state for efficient reweighted score matching. By conducting distributed online RL training across heterogeneous embodiments, the resulting fine-tuned policy, X-NavDP, achieves state-of-the-art cross-embodiment visual navigation performance, improving the overall success rate from 61.20\% to 84.28\% in simulation and 10\% to 65\% in real-world hard cases. The code and model are publicly available at \href{https://yty-sky.github.io/x-navdp-project-page/}{X-NavDP: Generalizing Navigation Diffusion Policy}.


     
\end{abstract}


\vspace{-0.1cm}

\vspace{-0.1cm}
\section{Introduction}
\vspace{-0.2cm}
Building general visual navigation policies upon large-scale simulation datasets has emerged as a promising research paradigm, offering excellent data scalability and strong zero-shot generalization across diverse indoor and outdoor scenes as well as heterogeneous robotic embodiments~\cite{navdp2024, peng2025logoplanner, zeng2025navidiffusor, wei2026navol, xie2025vid2sim, dualnav2024, ren2025prior}. Most approaches adopt imitation learning to optimize diffusion-based navigation policies, where policy behaviors are supervised by expert trajectories generated from an oracle global planner. Nevertheless, these privileged global trajectories are inaccessible to physical robots during real-world deployment, which creates a critical domain gap between offline training and online inference. Such imitation learning pipelines suffer from two inherent drawbacks for visual navigation tasks.

First, mimicking globally optimal trajectories inevitably induces decision ambiguity, since robots can only obtain partial local visual observations for real-time decision-making. The inherent mismatch between global offline supervision and local online observations severely suppresses policy exploration and autonomous failure recovery capabilities, leading to degraded performance in tough scenarios such as dead-end escape and long obstacle detour. Second, the embodiment-blind data generation pipeline neglects robot-specific dynamics and physical motion constraints, resulting in unstable and inconsistent navigation performance when transferring pretrained policies across different robot platforms.

To address the aforementioned drawbacks of pure imitation learning, reinforcement learning (RL) serves as a feasible refinement paradigm. It allows navigation policies to conduct interactive trial-and-error learning in different environments, and further adapt to local observation constraints and unique robot dynamics via self-collected sequential navigation experience~\cite{intelligence2025pi, yu2025rlinf, sheng2026beyond, xu2026rl, chen2025conrft}. Despite these merits, fine-tuning pretrained diffusion navigation policies via online RL still faces non-negligible technical bottlenecks. Specifically, policy gradient-based diffusion fine-tuning methods suffer from severe training instability, caused by chained sequential likelihood calculation throughout the full diffusion denoising process~\cite{black2024ddpo, dppo2024}. In contrast, latent space-based methods freeze core diffusion parameters during optimization, which severely limits policy exploration and fails to generate flexible detour and backward trajectories required for long-horizon complex navigation~\cite{wagenmaker2025steering}. As a lightweight and stable alternative, score-based reweighted methods~\cite{dpmd2024} can preserve pretrained navigation priors and avoid catastrophic forgetting and mode collapse during RL fine-tuning. However, it remains largely under-explored to tailor such reweighted mechanisms for high-dimensional visual inputs and long-horizon sequential navigation decision-making tasks.

To that end, we propose a novel Group Q-score Reweighted Matching (GQRM) framework equipped with two complementary designs to enable efficient and stable online RL fine-tuning for diffusion navigation policies. First, to diversify navigation behaviors while retaining inherent diffusion priors, we propose a self-bootstrapped trajectory perturbation module, which replaces the raw noise injection adopted in previous diffusion RL methods. Second, to compensate for insufficient learning signals for sparse and rarely visited hard navigation cases, we introduce within-group normalized Q-scores as adaptive reweighted coefficients, instead of vanilla globally normalized Q-scores that lack sensitivity to hard low-return states. Fig.~\ref{fig:teaser} provides a high-level overview.

Integrated with our proposed GQRM strategy, the post-trained diffusion navigation policy, X-NavDP achieves consistent cross-embodiment generalization with only marginal extra parameters introduced. Extensive experiments in both simulated environments and real-world robotic scenarios demonstrate that our method achieves superior state-of-the-art navigation performance. Specifically, our approach improves navigation success rate from 61.20\% to 84.28\% and SPL metric from 58.95\% to 77.19\% in simulation, and surpasses the previous approach in complex real-world layouts from 10\% to 65\%, consuming only 12 hours for post-training.

\vspace{-0.2cm}






\section{Related Work}
\vspace{-0.2cm}
\label{sec:related}

\subsection{Diffusion Models for Robot Navigation}
\vspace{-0.1cm}
Diffusion models~\citep{ho2020ddpm, song2021scorebased} generate samples through iterative denoising and are particularly effective at modeling multimodal data distributions. Diffusion Policy~\citep{chi2023diffusionpolicy} demonstrated the effectiveness of this generative formulation for robotic manipulation, with subsequent studies extending diffusion-based policies to more diverse and challenging manipulation tasks~\cite{black2024pi0, ze20243d, ke20243d}. Building on these advances, NoMaD~\citep{sridhar2024nomad} introduced diffusion-based trajectory generation for visual navigation and showed that multimodal prediction improves image-goal navigation and open-world exploration. NavDP~\citep{navdp2024} further developed this paradigm through a generate-then-filter pipeline that achieves strong navigation performance using simulation-only training data. Despite these advances, existing methods rely primarily on imitation learning and are therefore limited by the coverage and quality of their demonstrations. In this work, we investigate how online interaction can further improve pretrained navigation policies while preserving their acquired capabilities.
\vspace{-0.1cm}
\subsection{RL Fine-Tuning of Diffusion Policies}
\vspace{-0.1cm}
Reinforcement learning provides a promising approach to continuously improving pretrained diffusion models through online interaction~\cite{intelligence2025pi, yu2025rlinf, li2025simplevla}. Representative policy-gradient methods include DDPO~\citep{black2024ddpo}, which models diffusion denoising as an MDP for image generation, and DPPO~\citep{dppo2024}, which extends this formulation to robotic control through a two-layer MDP coupling environment interaction with diffusion denoising. Alternative approaches avoid direct likelihood-based policy optimization. DPMD~\citep{dpmd2024} performs online policy improvement through reweighted score matching without explicitly sampling from the intractable target policy, whereas DSRL~\citep{dsrl2024} optimizes the initial diffusion noise, which may restrict the emergence of behaviors beyond the pretrained policy distribution. The work most closely related to ours is that of \citet{sheng2026beyond}, which also applies online RL to fine-tune NavDP and demonstrates the benefits of interactive experience. However, its exploration relies solely on the stochasticity inherent in diffusion sampling, without an explicit mechanism for generating behaviorally diverse trajectories. In contrast, our method combines structured trajectory perturbation, embodiment-conditioned modulation, and within-group Q-score reweighting for efficient exploration and stable policy improvement.

\vspace{-0.1cm}
\section{Preliminaries}
\label{sec:prelim}

\vspace{-0.1cm}
\subsection{Diffusion Policy for Navigation}
\vspace{-0.1cm}
Following the prior works~\cite{sridhar2024nomad, navdp2024}, we formulate visual navigation as a conditional trajectory generation problem. Given an observation $\mathbf{o}_t$ consisting of RGB-D images and a goal specification $\mathbf{g}$, the policy generates an action chunk $a^{0} = \{(x_i, y_i)\}_{i=1}^{H}$ containing $H$ future waypoints through iterative denoising. The forward diffusion process adds Gaussian noise over $K$ steps:
\begin{equation}
    q(a^k | a^{k-1}) = \mathcal{N}(a^k; \sqrt{1-\beta_k}a^{k-1}, \beta_k \mathbf{I}),
\end{equation}
where $\beta_k$ is the noise schedule. The reverse process learns a noise prediction network $\boldsymbol{\epsilon}_\phi(a^k, k, \mathbf{o}_t, \mathbf{g})$ trained with mean squared error (MSE) loss:
\begin{equation}
    \mathcal{L}_{\text{dp}} = \mathbb{E}_{k, \boldsymbol{\epsilon}} \left[ w(k) \| \boldsymbol{\epsilon} - \boldsymbol{\epsilon}_\phi(a^k, k, \mathbf{o}_t, \mathbf{g}) \|^2 \right],
\end{equation}
where $w(k)$ is a timestep-dependent weight. At inference, action chunks are generated by iteratively denoising from $a^K \sim \mathcal{N}(\mathbf{0}, \mathbf{I})$ using the learned reverse process.
\vspace{-0.1cm}
\subsection{Policy Mirror Descent and Diffusion Policy Mirror Descent}
\vspace{-0.1cm}
\label{sec:pmd}

Policy mirror descent (PMD) is closely related to trust-region policy optimization methods such as TRPO~\citep{schulman2015trpo} and PPO~\citep{schulman2017ppo}. Rather than imposing a hard trust-region constraint, PMD regularizes each policy update through a KL-divergence proximal term~\citep{tomar2021mirror, lan2023policy}. Given the current policy $\pi_{\mathrm{old}}$, the policy-improvement target for each state $s$ is defined as
\begin{equation}
    \pi_{\mathrm{MD}}(\cdot|s)
    =
    \arg\max_{\pi(\cdot|s)}
    \;
    \mathbb{E}_{a\sim\pi(\cdot|s)}
    \left[Q^{\pi_{\mathrm{old}}}(s,a)\right]
    -
    \lambda
    D_{\mathrm{KL}}\!\left(
        \pi(\cdot|s)
        \,\|\,
        \pi_{\mathrm{old}}(\cdot|s)
    \right),
    \label{eq:pmd}
\end{equation}
where $Q^{\pi_{\mathrm{old}}}(s,a)$ is the state-action value function under $\pi_{\mathrm{old}}$, and $\lambda$ controls the strength of the KL regularization. The resulting policy has the closed-form solution
\begin{equation}
    \pi_{\mathrm{MD}}(a|s)
    =
    \frac{
        \pi_{\mathrm{old}}(a|s)
        \exp\!\left(Q^{\pi_{\mathrm{old}}}(s,a)/\lambda\right)
    }{
        Z_{\mathrm{MD}}(s)
    },
    \label{eq:pmd_solution}
\end{equation}
with partition function $Z_{\mathrm{MD}}(s)=\int\pi_{\mathrm{old}}(a|s)\exp\!\left(Q(s,a)/\lambda\right)da$.

\paragraph{Diffusion Policy Mirror Descent (DPMD).}
When the policy is parameterized by a diffusion model, directly sampling from the mirror-descent target $\pi_{\mathrm{MD}}(\cdot|s)$ is generally intractable because its density involves the unknown partition function $Z_{\mathrm{MD}}(s)$. DPMD~\citep{dpmd2024} addresses this problem through reweighted score matching. Let $p_t(a_t|s)=\int q_{t|0}(a_t|a_0)\pi_{\mathrm{MD}}(a_0|s)\,da_0$ denote the noised marginal distribution obtained by applying the forward diffusion process to the mirror-descent target. DPMD defines the reweighting function
\begin{equation}
    g_{\mathrm{MD}}(a_t;s)
    =
    Z_{\mathrm{MD}}(s)\,p_t(a_t|s),
    \label{eq:dpmd_reweighting}
\end{equation}
which scales the noised target marginal distribution by the state-dependent partition function. The corresponding reweighted score-matching objective is
\begin{align}
    \mathcal{L}_{\mathrm{DPMD}}(\theta;s,t)
    &=
    \int
    g_{\mathrm{MD}}(a_t;s)
    \left\|
        s_\theta(a_t;s,t)
        -
        \nabla_{a_t}\log p_t(a_t|s)
    \right\|^2
    da_t
    \nonumber\\
    &=
    \mathbb{E}_{
        \substack{
            a_0\sim\pi_{\mathrm{old}}(\cdot|s)\\
            a_t\sim q_{t|0}(\cdot|a_0)
        }
    }
    \left[
        \exp\!\left(
            \frac{Q^{\pi_{\mathrm{old}}}(s,a_0)}{\lambda}
        \right)
        \left\|
            s_\theta(a_t;s,t)
            -
            \nabla_{a_t}\log q_{t|0}(a_t|a_0)
        \right\|^2
    \right].
    \label{eq:dpmd}
\end{align}
This objective can be optimized using samples from the current policy. Specifically, a denoised action $a_0$ is sampled from $\pi_{\mathrm{old}}(\cdot|s)$ and noised to diffusion timestep $t$, after which the score network is regressed toward the conditional score $\nabla_{a_t}\log q_{t|0}(a_t|a_0)$. The exponential value term assigns larger regression weights to higher-value actions, enabling policy improvement in score-function space without requiring direct samples from $\pi_{\mathrm{MD}}$. Our GQRM objective in Eq.~\ref{eq:weighted_rsm} further improves this formulation by introducing within-group normalized Q-scores, as described in Sec.~\ref{sec:gqrm}.


\vspace{-0.2cm}
\section{Method}
\vspace{-0.1cm}
\label{sec:method}

\begin{figure}[t]
    \centering
    \includegraphics[width=0.90\linewidth]{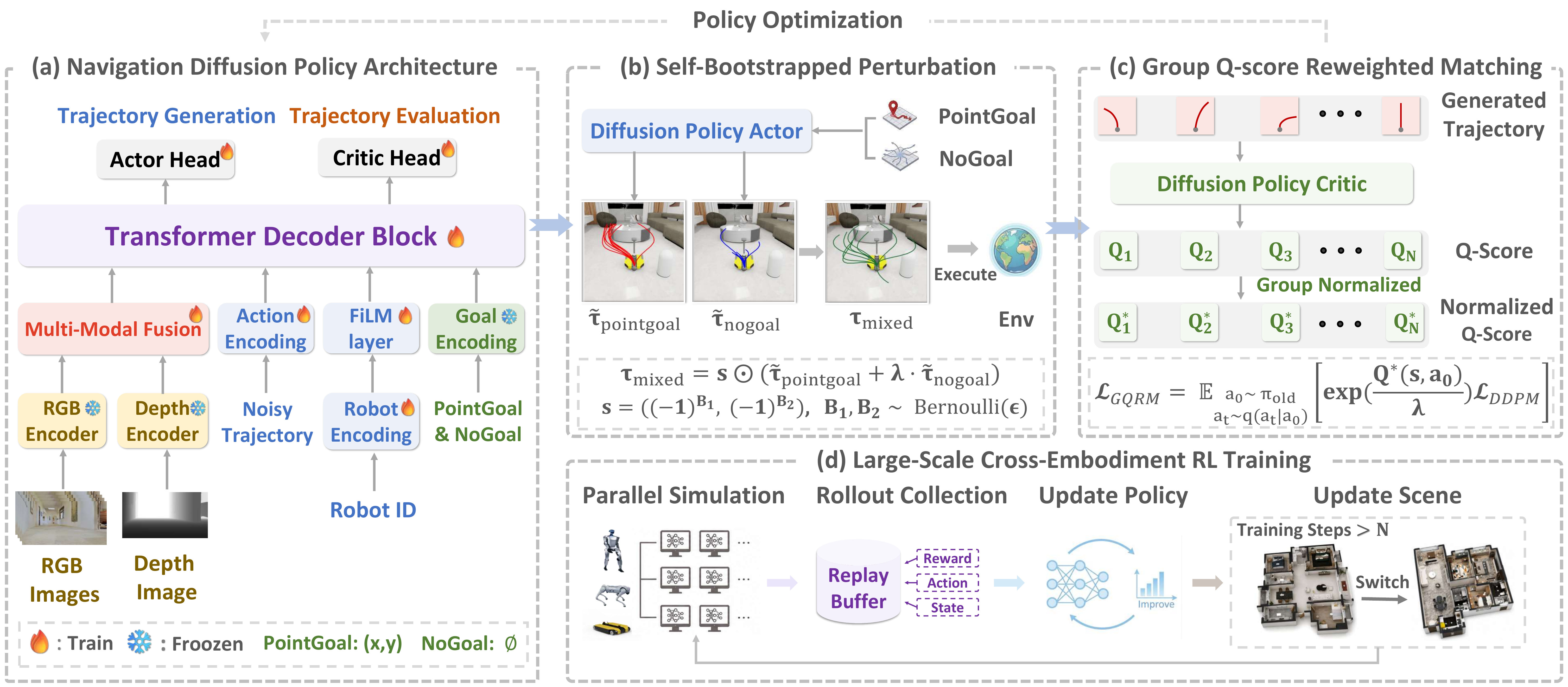}
    \caption{Overview of the \ours{} pipeline. (a) We adopt the NavDP backbone and inject embodiment information via FiLM layers. (b) A structured trajectory perturbation scheme mixes goal-conditioned and goal-agnostic samples for efficient exploration (denoted as pointgoal and nogoal in the figure, respectively). (c) Within-group $Q$-score normalization produces adaptive weights for reweighted score matching. (d) The framework enables large-scale cross-embodiment RL.}
    \label{fig:pipeline}
    \vspace{-0.4cm}
\end{figure}

\ours{} is an RL post-trained diffusion policy for navigation, built upon the pretrained NavDP~\citep{navdp2024}.  Fig.~\ref{fig:pipeline} shows the overall pipeline. Given local visual observations and navigation goals, our policy outputs an action chunk with $H$-step future waypoints. Our core objective is to enhance policy performance through online interactive simulation experience, while preserving the generalization ability of the pretrained model. To achieve this, we develop three complementary modules: (1) self-bootstrapped perturbation for efficient exploration, (2) group Q-score reweighted matching (GQRM) for stable policy iteration, and (3) embodiment-conditioned modulation for cross-embodiment policy training. We elaborate on these three components in the following subsections.




\vspace{-0.1cm}
\subsection{RL Post-Training Framework}
\vspace{-0.1cm}

\label{sec:pipeline}
\label{sec:training}

Massive parallelization is essential for data-efficient RL training~\cite{rudin2022learning, mayor2025impact}. We build a unified RL post-training framework on IsaacLab~\cite{mittal2025isaac}, where over $500$ wheeled, quadruped, and humanoid robots can interact with the environments simultaneously across $50+$ scenes. All rollouts are aggregated into a shared buffer for joint cross-embodiment policy optimization. Because the navigation policy outputs waypoint chunks instead of joint commands, we employ a hierarchical control stack: the diffusion policy predicts a short-horizon target trajectory, a unified MPC converts it into base velocity commands, and an embodiment-specific pretrained locomotion policy executes the commands. The MPC and locomotion controllers run at $25$\,Hz, while the navigation policy is invoked only chunk-wise: each predicted trajectory is tracked for $3$ seconds, and the cumulative reward over this window is stored as one macro-step transition. This protocol reduces inference overhead and aligns value learning with trajectory-level actions.

\vspace{-0.15cm}
\subsection{Self-Bootstrapped Perturbation}
\vspace{-0.15cm}
\label{sec:nogoal}


A pretrained goal-conditioned diffusion policy carries a strong navigation prior, so its goal-conditioned samples concentrate on a narrow set of forward-moving trajectories toward the point goal. Online exploration is therefore overly conservative, especially in hard states requiring backing out, lateral motion, or short detours. The resulting rollouts rarely contain the recovery behaviors that RL needs to reinforce. A natural remedy is to inject Gaussian noise into the trajectory, but unstructured noise seldom induces meaningful navigation behavior and easily destroys the temporal smoothness and dynamic feasibility. Our insight is to leverage the fact that the same pretrained navigation policy, when conditioned on a goal-agnostic condition, produces scene-consistent trajectories yet explores more aggressively than its goal-conditioned counterpart. We therefore use this internal goal-agnostic branch to perturb goal-conditioned trajectories, increasing behavioral diversity while preventing excessive deviation from the pretrained trajectory manifold.

For each observation, we draw a goal-conditioned trajectory $\tilde{\boldsymbol{\tau}}_{\text{pointgoal}}$ and a goal-agnostic trajectory $\tilde{\boldsymbol{\tau}}_{\text{nogoal}}$ from the same visual context, and construct the mixed trajectory as:
\begin{gather}
    \boldsymbol{\tau}_{\text{mixed}}
    \;=\;
    \mathbf{s}\odot
    \bigl(
    \tilde{\boldsymbol{\tau}}_{\text{pointgoal}}
    \;+\;
    \lambda\,\tilde{\boldsymbol{\tau}}_{\text{nogoal}}
    \bigr),
    \quad
    \mathbf{s}=\bigl((-1)^{B_1},(-1)^{B_2}\bigr)\\
    B_1, B_2\sim\operatorname{Bernoulli}(\epsilon),
    \quad
    B_1\perp B_2,
    \label{eq:nogoal_mix}
\end{gather}
where $\lambda$ is a signed mixing coefficient and the Hadamard product $\odot$ applies the sign vector $\mathbf{s}$ to every waypoint, independently flipping the $x$ and $y$ coordinates of the entire chunk with probabilities $\epsilon$. This enables the policy to produce lateral, backward-recovery, and detour trajectories while preserving the smooth, dynamically feasible structure of the original waypoints. The same perturbed set is used for both rollout collection and actor updates, keeping the data-collection and policy-improvement distributions aligned.

\vspace{-0.1cm}
\subsection{GQRM: Group Q-score Reweighted Matching}
\vspace{-0.1cm}
\label{sec:gqrm}

GQRM turns the self-bootstrapped candidate group into a stable diffusion actor update. Following DPMD~\citep{dpmd2024}, a diffusion actor can be optimized by reweighted score matching, sidestepping the likelihood-ratio gradients needed for back-propagation through the long reverse chain. However, the original Q-weighted variant in DPMD becomes uninformative in hard states where all sampled candidates share similarly low absolute $Q$-scores: DPMD normalizes Q-scores and computes the exponential weights across a minibatch, which mixes candidates from unrelated states, causing easy states with higher absolute returns to dominate the actor update and failing to provide meaningful gradients for learning recovery actions in hard states. We therefore normalize values \emph{within each same-state candidate group} drawn from $\pi_{\mathrm{old}}(\cdot|s)$, rather than across unrelated states in a minibatch. This ensures that even when absolute returns are low, the actor can still shift probability mass toward locally better candidates.

For a denoised action $a_0\sim\pi_{\mathrm{old}}(\cdot|s)$---where $a_0$ denotes an $H$-step trajectory chunk---we define the within-group statistics
\begin{equation}
    \bar Q_G(s)
    \;=\;
    \mathbb{E}_{a_0\sim\pi_{\mathrm{old}}(\cdot|s)}\!\bigl[Q(s,a_0)\bigr],
    \qquad
    \sigma_G(s)
    \;=\;
    \sqrt{\mathbb{E}_{a_0\sim\pi_{\mathrm{old}}(\cdot|s)}\!\bigl[(Q(s,a_0)-\bar Q_G(s))^2\bigr]}.
    \label{eq:q_weights}
\end{equation}
The group-normalized value used in the actor weight is
\begin{equation}
    \widetilde{Q}_G(s,a_0) = \mathrm{clip}\!\left(\frac{c\,\bigl(Q(s,a_0)-\bar Q_G(s)\bigr)}{\sigma_G(s)+\varepsilon},\;-h,\;h\right),
    \label{eq:group_q}
\end{equation}
where $c$ scales the normalized advantage, $h$ caps its magnitude, and a small $\varepsilon>0$ guards against degenerate groups. Let $a_t\sim q_{t|0}(\cdot|a_0)$ denote the noised action at diffusion timestep $t$. The GQRM objective is
\begin{align}
    \mathcal{L}_{\text{GQRM}}(\theta;\,s,t)
    \;=\;
    \mathbb{E}_{\substack{a_0\sim\pi_{\mathrm{old}}(\cdot|s)\\ a_t\sim q_{t|0}(\cdot|a_0)}}
    \!\left[
        \exp\!\bigl(\widetilde{Q}_G(s,a_0)/\lambda\bigr)\;
        \bigl\|
            s_\theta(a_t;\,s,t) \;-\; \nabla_{a_t}\log q_{t|0}(a_t|a_0)
        \bigr\|^2
    \right],
    \label{eq:weighted_rsm}
\end{align}
where $\lambda$ is a temperature parameter controlling the peakedness of the actor weights; the detailed derivation is deferred to Appendix~\ref{app:gqrm_derivation}. In practice, the expectations over $\pi_{\mathrm{old}}(\cdot|s)$ are estimated by the same-state candidate group $\mathcal{G}(s)$ produced by self-bootstrapped perturbation (Sec.~\ref{sec:nogoal}), and we retain only the top-$k$ candidates with positive advantage $\widetilde{Q}_G(s,a_0)>0$, suppressing noise from low-value samples and reducing compute.

\subsection{Cross-Embodiment Training and Closed-Loop Execution}
\label{sec:film}
\label{sec:rtc}

Architecturally, we modify the original NavDP model by adding an embodiment-modulated module that injects a learned robot embedding at two complementary points within the diffusion decoder. The noisy waypoint chunk is first lifted into action-token features $\mathbf{u}$ through the action embedding layer. Given an embodiment index $e$, a learned robot embedding $\mathbf{z}_e = E_{\mathrm{emb}}(e)$ shifts every action token before decoding,
\begin{equation}
    \mathbf{u}_e \;=\; \mathbf{u} + f_{\Delta}(\mathbf{z}_e).
\end{equation}
The decoder then fuses the embodiment-modulated action tokens with the visual--goal conditioning memory into hidden trajectory features $\mathbf{h}$. The same robot embedding is further used to generate FiLM~\citep{perez2018film} modulation parameters after the decoder,
\begin{equation}
    [\Delta\boldsymbol{\gamma}_e,\Delta\boldsymbol{\beta}_e]
    \;=\;
    f_{\mathrm{FiLM}}(\mathbf{z}_e),
    \qquad
    \mathbf{h}_e \;=\; (1+\Delta\boldsymbol{\gamma}_e)\odot\mathbf{h} + \Delta\boldsymbol{\beta}_e,
    \qquad
    \hat{\boldsymbol{\epsilon}}_\phi \;=\; H_{\mathrm{act}}(\mathbf{h}_e),
    \label{eq:film}
\end{equation}
where $H_{\mathrm{act}}$ is the denoising head. The embedding therefore modulates both the action-token pathway before the decoder and the trajectory-feature pathway after the decoder, allowing a single shared noise predictor to generate embodiment-specific trajectories without replicating policy weights across robots.
\paragraph{Closed-Loop Temporal Guidance.}
During deployment, we add a lightweight temporal-consistency term inspired by Real-Time Chunking (RTC)~\citep{black2026real} to reduce discontinuities between consecutive receding-horizon predictions. At execution step $t$, let $\mathbf{y}_t$ denote the previously committed trajectory transformed into the current robot frame, and let $\mathbf{w}_t$ be a vector of decaying prefix weights that emphasizes near-term consistency while allowing the far horizon to replan. During DDPM sampling, let $\mathbf{x}_t^{(k)}$ denote the noisy trajectory at reverse step $k$, and let $\hat{\mathbf{x}}_t^{(0,k)}$ denote the one-step estimate of the denoised trajectory $\mathbf{x}_0$ predicted from $\mathbf{x}_t^{(k)}$ at that reverse step. We form the guidance gradient
\begin{equation}
    \mathbf{g}
    \;=\;
    \nabla_{\mathbf{x}_t^{(k)}}
    \bigl\langle
        \operatorname{sg}\!\bigl(\mathbf{w}_t\odot(\mathbf{y}_t-\hat{\mathbf{x}}_t^{(0,k)})\bigr),\;
        \hat{\mathbf{x}}_t^{(0,k)}
    \bigr\rangle,
    \label{eq:rtc_grad}
\end{equation}
where $\odot$ is the Hadamard product, $\operatorname{sg}(\cdot)$ stops gradient flow through the residual term, and $\langle\cdot,\cdot\rangle$ accumulates the weighted consistency residual across waypoint dimensions and horizon indices. The resulting vector--Jacobian product encourages the prediction to move closer to the previously committed trajectory without overriding the learned diffusion prior. The guided reverse step becomes
\begin{equation}
    \mathbf{x}_t^{(k-1)}
    \;=\;
    \boldsymbol{\mu}_k \;+\; \sigma_k\,\mathbf{z}
    \;+\; \sqrt{\bar{\alpha}_k}\;\eta_{\text{guide}}\;\mathbf{g},
    \label{eq:rtc_step}
\end{equation}
where $\boldsymbol{\mu}_k$ and $\sigma_k$ are the DDPM posterior mean and standard deviation at reverse step $k$, $\mathbf{z}\sim\mathcal{N}(\mathbf{0},\mathbf{I})$, $\bar{\alpha}_k$ is the cumulative noise-schedule coefficient, and $\eta_{\text{guide}}$ controls the guidance intensity.


\vspace{-0.2cm}
\section{Experiments}
\vspace{-0.1cm}
\label{sec:experiments}

We design experiments to answer four questions: (1)~Does \ours{} outperform strong navigation baselines across different embodiments and scenes? (2)~Does RL post-training induce new navigation behaviors such as recovery and replanning? (3)~How much does each proposed component contribute to the final performance? (4)~What failure modes remain?

\subsection{Experimental Setup}
\label{sec:setup}

We build our simulation benchmark in IsaacLab~\citep{mittal2023orbit} using scenes from GRScenes-100~\cite{InternScenes, internnav2025}. For RL post-training, we use 56 training scenes, including 47 home scenes and 9 commercial scenes. For held-out evaluation, we use 40 unseen scenes, including 20 home scenes and 20 commercial scenes. All scenes are loaded as USD assets with realistic collision geometry and physics simulation. \emph{Simulation:} we evaluate three embodiments: \textbf{Dingo} (differential-drive wheeled), \textbf{Unitree~Go2} (quadruped), and \textbf{Unitree~G1} (humanoid). \emph{Real hardware:} we set up lab, hall, and office test environments and deploy the same simulation-trained policy on \textbf{Turtlebot}, \textbf{Unitree~Go2}, and \textbf{Unitree~G1} with onboard RGB-D cameras. Each real-world setting is evaluated over 10 trials.

We report two standard navigation metrics: \textbf{Success Rate (SR~$\uparrow$)}, the percentage of episodes reaching within $0.5$\,m of the goal, and \textbf{SPL~($\uparrow$)}, success weighted by path length~\citep{anderson2018evaluation}, which jointly reflects success and path efficiency. We compare against navigation baselines iPlanner~\cite{yang2023iplanner}, ViPlanner~\cite{roth2024viplanner}, NavDP~\citep{navdp2024}, NavOL~\cite{wei2026navol}, SIDP~\cite{zhang2026self}, and NavDP-RL~\cite{sheng2026beyond}. 



\vspace{-0.2cm}
\subsection{Main Results}
\vspace{-0.1cm}
\label{sec:main_results}

\begin{figure}[t]
    \centering
    \includegraphics[width=0.85\linewidth]{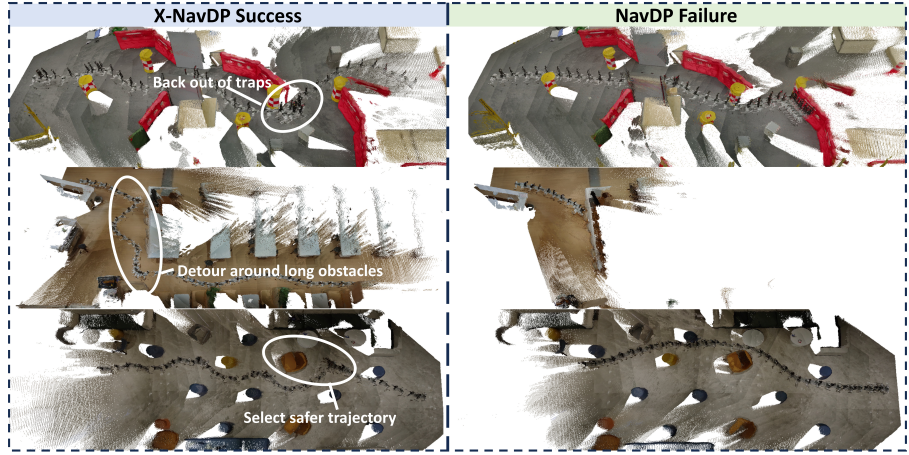}
    \caption{Qualitative trajectory comparison. \ours{} produces recovery, detour, and safer behaviors that are absent from the NavDP policy. The visualization is provided with VGGT-$\Omega$~\cite{wang2026vggt}.}
    \label{fig:qualitative}
    \vspace{-0.4cm}
\end{figure}

Tables~\ref{tab:main_sim} and~\ref{tab:main_real} compare \ours{} with navigation baselines in simulation and real-world settings. In simulation, \ours{} consistently outperforms the NavDP base policy across all embodiments and scenes, improving the average SR from 61.20\% to 84.28\% and SPL from 58.95\% to 77.19\%. The gains are especially pronounced for humanoid robots: while the vanilla pretrained policy struggles in both commercial and home scenes, \ours{} raises its SR from 64.35\%/50.70\% to 84.25\%/84.50\%. The consistent improvements across embodiments and scenes indicate strong cross-embodiment generalization. Qualitatively, Fig.~\ref{fig:qualitative} shows that the fine-tuned policy learns new capabilities beyond imitation, including backing out of traps, detouring around long obstacles, and selecting safer paths when multiple navigation options are available.

\begin{table}[t]
\centering
\caption{Point-goal navigation results in IsaacLab across 40 indoor scenes. Each cell reports SR/SPL. Best values are in \textbf{bold}.}
\label{tab:main_sim}
\scriptsize
\begin{tabular}{l c c c c c c c}
\toprule
& Overall & \multicolumn{2}{c}{Wheeled} & \multicolumn{2}{c}{Quadruped} & \multicolumn{2}{c}{Humanoid} \\
\cmidrule(lr){3-4} \cmidrule(lr){5-6} \cmidrule(lr){7-8}
Method & SR/SPL$\uparrow$ & Commercial & Home & Commercial & Home & Commercial & Home \\
\midrule
iPlanner~\cite{yang2023iplanner} & 33.84/32.60 & 51.75/50.02 & 38.80/37.05 & 53.05/51.94 & 42.60/41.44 & 10.90/10.23 & 5.95/4.94 \\
ViPlanner~\cite{roth2024viplanner} & 43.87/42.84 & 55.40/53.94 & 34.75/34.05 & 55.85/54.32 & 40.09/38.35 & 46.90/46.36 & 30.25/30.02 \\
NavDP~\cite{navdp2024} & 61.20/58.95 & 72.55/69.74 & 68.10/63.96 & 58.10/56.86 & 53.45/51.16 & 64.35/63.15 & 50.70/48.84 \\
NavOL~\cite{wei2026navol} & -- & 75.20/71.10 & 58.80/54.90 & -- & -- & -- & -- \\
SIDP~\cite{zhang2026self} & -- & 81.19/73.36 & 63.17/56.48 & -- & -- & -- & -- \\
NavDP-RL~\cite{sheng2026beyond} & -- & -- & 68.70/62.80 & -- & -- & -- & -- \\
\midrule
\textbf{\ours{}} & \textbf{84.28/77.19} & \textbf{88.55/78.96} & \textbf{88.70/76.37} & \textbf{80.65/75.72} & \textbf{79.05/72.35} & \textbf{84.25/80.41} & \textbf{84.50/79.37} \\
\bottomrule
\end{tabular}
\vspace{-0.5cm}
\end{table}

\begin{table}[t]
\centering
\caption{Real-world point-goal deployment. Policy weights are trained in simulation and deployed without sim-to-real fine-tuning.}
\label{tab:main_real}
\scriptsize
\begin{tabular}{l ccc ccc ccc}
\toprule
& \multicolumn{3}{c}{Wheeled} & \multicolumn{3}{c}{Quadruped} & \multicolumn{3}{c}{Humanoid} \\
\cmidrule(lr){2-4} \cmidrule(lr){5-7} \cmidrule(lr){8-10}
Method & Lab & Hall & Office & Lab & Hall & Office & Lab & Hall & Office \\
\midrule
iPlanner & 0.00 & 20.0 & 0.00 & 0.00 & 20.0 & 0.00 & 0.00 & 0.00 & 0.00 \\
ViPlanner & 0.00 & 50.0 & 0.00 & 0.00 & 40.0 & 0.00 & 0.00 & 30.0 & 0.00 \\
NavDP~\citep{navdp2024} & 0.00 & 40.0 & 0.00 & 0.00 & 30.0 & 0.00 & 0.00 & 20.0 & 0.00 \\
\textbf{\ours{}} & \textbf{60.0} & \textbf{70.0} & \textbf{60.0} & \textbf{80.0} & \textbf{60.0} & \textbf{70.0} & \textbf{50.0} & \textbf{50.0} & \textbf{80.0} \\
\bottomrule
\end{tabular}
\vspace{-0.5cm}
\end{table}

\vspace{-0.5cm}
\begin{table}[!h]
\centering
\caption{RL method comparison on NavDP in the 24-scene setting. Each cell reports SR/SPL.}
\label{tab:ablation_rl}
\scriptsize
\resizebox{\textwidth}{!}{
\begin{tabular}{l c c c c c c c}
\toprule
& Overall & \multicolumn{2}{c}{Wheeled} & \multicolumn{2}{c}{Quadruped} & \multicolumn{2}{c}{Humanoid} \\
\cmidrule(lr){3-4} \cmidrule(lr){5-6} \cmidrule(lr){7-8}
Method & SR/SPL$\uparrow$ & Commercial & Home & Commercial & Home & Commercial & Home \\
\midrule
SFT & 53.50/50.84 & 73.44/69.74 & 55.05/51.59 & 52.00/49.59 & 36.63/34.78 & 57.90/55.46 & 45.99/ 43.88 \\
DPPO~\citep{dppo2024} & 33.98/33.28 & 50.00/49.20 & 27.80/26.52 & 32.10/31.73 & 18.40/17.91 & 55.80/54.90 & 19.75/19.42 \\
DSRL~\citep{dsrl2024} & 13.65/12.26 & 11.80/10.82 & 7.35/6.57 & 17.35/15.54 & 14.25/12.78 & 18.75/16.62 & 12.40/11.28 \\
DPMD-original (w/o group)~\citep{dpmd2024} & 71.57/67.32 & 79.55/73.10 & 73.85/66.35 & 75.45/71.46 & 67.30/62.92 & 71.65/70.27 & 61.65/59.84 \\
\textbf{\ours{}} & \textbf{80.48/74.78} & \textbf{85.80/79.48} & \textbf{85.75/76.49} & \textbf{78.45/73.66} & \textbf{74.25/68.69} & \textbf{81.85/78.13} & \textbf{76.75/72.25} \\
\bottomrule
\end{tabular}
}
\vspace{-0.2cm}
\end{table}


\begin{figure}[!h]
    \centering
    \includegraphics[width=0.88\linewidth]{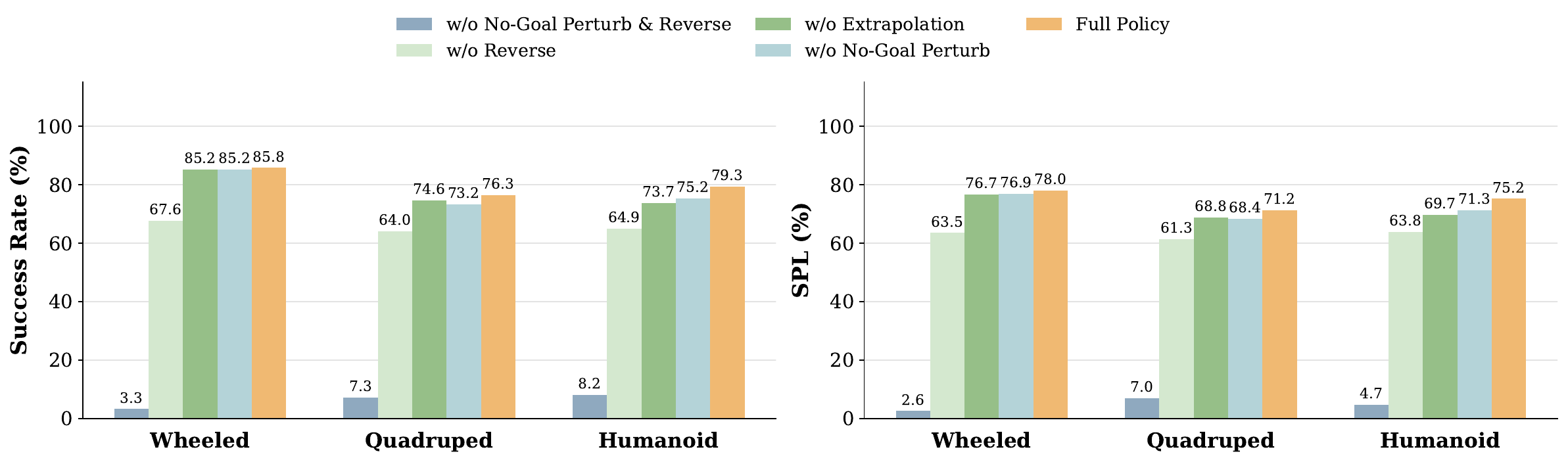}
    \caption{Ablation study on exploration policies. The results demonstrate that no-goal perturbation and trajectory reversal are two critical designs for policy learning. }
    \label{fig:ablation}
    \vspace{-0.5cm}
\end{figure}
\vspace{-0.2cm}
\subsection{Ablation Studies}
\vspace{-0.2cm}

\label{sec:ablation}

To improve ablation efficiency, we conduct all ablation experiments on a 24-scene subset consisting of 6 commercial scenes and 18 home scenes, which is 32 fewer scenes than the main experiments.
\vspace{-0.2cm}
\paragraph{RL Fine-Tuning Method Comparison.}
\label{sec:ablation_rl}

Table~\ref{tab:ablation_rl} analyzes the impact of our post-training objective. SFT denotes supervised fine-tuning on trajectory data collected from the three embodiments (Dingo, Unitree Go2, and Unitree G1), without RL reweighting. Policy-gradient fine-tuning is unstable for diffusion navigation policies, with DPPO producing chaotic trajectories that degrade performance. DSRL collapses almost entirely, indicating that optimizing only the initial noise is insufficient for learning effective navigation behaviors. DPMD-style reweighted score matching provides a stronger baseline than DPPO and DSRL, while \ours{} further improves performance through same-state group normalization. This shows that same-state $Q$-score normalization is important, especially for efficient learning in hard states.

\vspace{-0.25cm}
\paragraph{Exploration Policy Ablation.}
\label{sec:ablation_nogoal}


Fig.~\ref{fig:ablation} shows that exploration quality is a key bottleneck for RL post-training. Removing both goal-agnostic perturbation and reverse exploration causes training to collapse, highlighting the need for diverse exploration in diffusion policy learning. Using either goal-agnostic perturbation or reverse exploration alone recovers substantial performance, while the full exploration policy achieves the best overall SR/SPL by generating more diverse candidate behaviors. We also evaluate a variant that combines goal-agnostic and point-goal trajectories with a fixed weight of 1. This variant underperforms our full perturbation strategy because it lacks extrapolation ability and remains constrained by the action manifold of the pretrained policy.

\vspace{-0.25cm}
\paragraph{Failure Case Analysis}
\label{sec:failure}
Our failure cases mainly arise from embodiment constraints, limited long-horizon memory, and imperfect obstacle perception. For G1, narrow passages are particularly challenging because the humanoid's bulky body and large turning radius make traversal difficult and can cause its arms to contact nearby obstacles while walking. Failures also occur during detours around long obstacles, especially when the goal lies on the opposite side of a wall. Although our method mitigates this issue, limited memory can make the policy lose trajectory consistency, causing it to return along its original path. The system also struggles with transparent and hollow obstacles, such as glass partitions and perforated panels, which are difficult to perceive from RGB images and can lead to unreliable depth measurements. These failure modes suggest that semantic obstacle understanding, stronger long-horizon memory, and stronger RGB perception could further improve performance.



\vspace{-0.4cm}
\section{Conclusion}
\vspace{-0.2cm}
\label{sec:conclusion}

We introduce \ours{}, an RL post-training framework for improving the general navigation ability of pretrained diffusion policies. Our core design combines goal-agnostic trajectory perturbation with group Q-score reweighted matching to enable structured exploration without dedicated curiosity modules. Lightweight embodiment modulation and RTC guidance further improve cross-robot generalization and temporal consistency. Experimental results show that RL post-training can substantially improve navigation diffusion policies while preserving their pretrained navigation priors.
\vspace{-0.45cm}
\paragraph{Limitations.} Our framework currently relies on short-term temporal context, which may limit its performance on tasks requiring long-term memory. It has been validated on three embodiments with predefined locomotion controllers, and adapting it to a new robot morphology requires a corresponding pretrained low-level controller. Although our real-world experiments cover both indoor and outdoor environments, broader evaluation across more diverse scenes, environmental conditions, and robot platforms remains an important direction for future work.
\vspace{-0.15cm}
\paragraph{Future Work.} We aim to scale the system through joint training over more diverse scenarios and richer robot morphologies. We also plan to investigate tighter coordination between decoupled locomotion and navigation policies for better overall navigation performance. In addition, we are interested in transferring this post-training framework to broader embodied navigation tasks, such as vision-language navigation.


\bibliography{example}

\clearpage
\appendix

\section{Group $Q$-score Reweighted Matching Derivation}
\label{app:gqrm_derivation}

We derive the GQRM actor objective by building on the reweighted score-matching framework of DPMD. Starting from the DPMD objective, GQRM replaces the raw Q-score weight with a same-state group-normalized Q-score weight. The key distinction is that each candidate group is constructed within a single state, so each state has its own normalization statistics and local policy-improvement target. We first revisit denoising score matching, then restate DPMD, and finally introduce the group-normalized reweighting step that leads to GQRM.

\paragraph{Notation.}
We use $a_0$ to denote the denoised action before the diffusion forward process and $a_t$ to denote its noised version at diffusion timestep $t$. In this paper, an ``action'' is an $H$-step trajectory chunk. The subscript in $a_0$ and $a_t$ denotes diffusion time and should not be confused with the candidate index. In the finite-sample implementation, $a_{0,i}$ denotes the $i$-th denoised candidate sampled from the same state. The subscript $G$ denotes quantities computed within the same-state candidate group, such as the group mean $\bar Q_G$, group standard deviation $\sigma_G$, and group-normalized value $\widetilde Q_G$.

\subsection{Score Matching Derivation for Diffusion Models}
We first derive the diffusion score-matching objective in our action notation. Let $p_0(a_0|s)$ be a denoised action distribution conditioned on state $s$, and let
\begin{equation}
    q_{t|0}(a_t|a_0)
    =
    \mathcal{N}\!\left(
        a_t;
        \sqrt{\bar{\alpha}_t}a_0,
        (1-\bar{\alpha}_t)\mathbf{I}
    \right)
\end{equation}
be the diffusion perturbation kernel at timestep $t$. The corresponding noised action distribution is
\begin{equation}
    p_t(a_t|s)
    =
    \int
    q_{t|0}(a_t|a_0)
    p_0(a_0|s)
    da_0 .
\end{equation}
Directly fitting the marginal score $\nabla_{a_t}\log p_t(a_t|s)$ is generally intractable because it requires integrating over all denoised actions. DDPM training therefore uses the tractable conditional Gaussian score as the denoising score-matching target:
\begin{equation}
    \mathcal{L}_{\mathrm{DSM}}(\theta)
    =
    \mathbb{E}_{t,\,
    a_0\sim p_0(\cdot|s),\,
    a_t\sim q_{t|0}(\cdot|a_0)}
    \left[
        \left\|
            s_\theta(a_t;s,t)
            -
            \nabla_{a_t}
            \log q_{t|0}(a_t|a_0)
        \right\|^2
    \right].
    \label{eq:app_dsm_objective}
\end{equation}
Under the population objective and sufficient model capacity, the minimizer of this denoising objective satisfies
\begin{equation}
    s_{\theta^*}(a_t;s,t)
    =
    \mathbb{E}_{a_0\sim q_{0|t}(\cdot|a_t,s)}
    \left[
        \nabla_{a_t}
        \log q_{t|0}(a_t|a_0)
    \right]
    =
    \nabla_{a_t}
    \log p_t(a_t|s).
    \label{eq:app_optimal_score}
\end{equation}

\emph{Proof.}
For a fixed timestep $t$, Tweedie's identity~\citep{efron2011tweedie} gives
\begin{equation}
    \nabla_{a_t}\log p_t(a_t|s)
    =
    \mathbb{E}_{a_0\sim q_{0|t}(\cdot|a_t,s)}
    \left[
        \nabla_{a_t}
        \log q_{t|0}(a_t|a_0)
    \right],
    \label{eq:app_tweedie}
\end{equation}
where
\begin{equation}
    q_{0|t}(a_0|a_t,s)
    =
    \frac{
        q_{t|0}(a_t|a_0)
        p_0(a_0|s)
    }{
        p_t(a_t|s)
    }.
\end{equation}
Indeed,
\begin{align}
    \nabla_{a_t}\log p_t(a_t|s)
    &=
    \frac{
        \nabla_{a_t}
        p_t(a_t|s)
    }{
        p_t(a_t|s)
    }
    \nonumber\\
    &=
    \frac{
        \nabla_{a_t}
        \int
        q_{t|0}(a_t|a_0)
        p_0(a_0|s)
        da_0
    }{
        p_t(a_t|s)
    }
    \nonumber\\
    &=
    \frac{
        \int
        \nabla_{a_t}
        \log q_{t|0}(a_t|a_0)
        q_{t|0}(a_t|a_0)
        p_0(a_0|s)
        da_0
    }{
        p_t(a_t|s)
    }
    \nonumber\\
    &=
    \mathbb{E}_{a_0\sim q_{0|t}(\cdot|a_t,s)}
    \left[
        \nabla_{a_t}
        \log q_{t|0}(a_t|a_0)
    \right].
    \label{eq:app_tweedie_verify}
\end{align}
We use the score network to match both sides of Eq.~\ref{eq:app_tweedie} and take the expected squared error under $a_t\sim p_t(\cdot|s)$. Matching the left-hand side is straightforward and gives the marginal score loss
\begin{equation}
    \mathbb{E}_{a_t\sim p_t(\cdot|s)}
    \left[
        \left\|
            s_\theta(a_t;s,t)
            -
            \nabla_{a_t}
            \log p_t(a_t|s)
        \right\|^2
    \right].
    \label{eq:app_marginal_score_loss}
\end{equation}
Matching the right-hand side gives an equivalent objective. Using Eq.~\ref{eq:app_tweedie}, the marginal score target can be replaced by the posterior expectation of the conditional Gaussian score:
\begin{align}
    &
    \mathbb{E}_{a_t\sim p_t(\cdot|s)}
    \left[
        \left\|
            s_\theta(a_t;s,t)
            -
            \mathbb{E}_{a_0\sim q_{0|t}(\cdot|a_t,s)}
            \nabla_{a_t}
            \log q_{t|0}(a_t|a_0)
        \right\|^2
    \right]
    \nonumber\\
    &=
    \mathbb{E}_{a_t\sim p_t}
    \left[
        \left\|
            s_\theta(a_t;s,t)
        \right\|^2
    \right]
    -
    2\,
    \mathbb{E}_{a_0\sim p_0,\,
    a_t\sim q_{t|0}}
    \left[
        \left\langle
            s_\theta(a_t;s,t),
            \nabla_{a_t}
            \log q_{t|0}(a_t|a_0)
        \right\rangle
    \right]
    + C_1
    \nonumber\\
    &=
    \mathbb{E}_{a_0\sim p_0(\cdot|s),\,
    a_t\sim q_{t|0}(\cdot|a_0)}
    \left[
        \left\|
            s_\theta(a_t;s,t)
            -
            \nabla_{a_t}
            \log q_{t|0}(a_t|a_0)
        \right\|^2
    \right]
    + C_2 ,
    \label{eq:app_conditional_score_loss}
\end{align}
where $C_1$ and $C_2$ are constants independent of $\theta$. The marginal score loss in Eq.~\ref{eq:app_marginal_score_loss} is minimized by $s_{\theta^*}(a_t;s,t)=\nabla_{a_t}\log p_t(a_t|s)$ over the $a_t$ space. Since Eq.~\ref{eq:app_conditional_score_loss} differs from this marginal objective only by a constant, it has the same optimal $\theta^*$. Therefore, the DSM objective in Eq.~\ref{eq:app_dsm_objective}, which optimizes this conditional denoising term, also yields the marginal score estimator in Eq.~\ref{eq:app_optimal_score}.

\subsection{Group Q-score Reweighted Matching}
Let $\pi_{\mathrm{old}}$ denote the behavior distribution induced by the current policy.

\paragraph{DPMD derivation.}
Following the PMD formulation in Sec.~\ref{sec:pmd}, DPMD constructs a KL-regularized policy-improvement target for each state $s$:
\begin{equation}
    \pi_{\mathrm{MD}}(\cdot|s)
    =
    \arg\max_{\pi}
    \;
    \mathbb{E}_{a_0\sim\pi(\cdot|s)}
    \big[Q(s,a_0)\big]
    -
    \lambda
    D_{\mathrm{KL}}\!\left(
        \pi(\cdot|s)\,\|\,\pi_{\mathrm{old}}(\cdot|s)
    \right).
    \label{eq:app_pmd_objective}
\end{equation}
Its closed-form solution is
\begin{equation}
    \pi_{\mathrm{MD}}(a_0|s)
    =
    \frac{
        \pi_{\mathrm{old}}(a_0|s)
        \exp\!\left(Q(s,a_0)/\lambda\right)
    }{
        Z_{\mathrm{MD}}(s)
    },
    \qquad
    Z_{\mathrm{MD}}(s)
    =
    \int
    \pi_{\mathrm{old}}(a_0|s)
    \exp\!\left(Q(s,a_0)/\lambda\right)
    da_0.
    \label{eq:app_md_target}
\end{equation}
DPMD uses $\pi_{\mathrm{MD}}(a_0|s)$ as the target distribution for policy learning. In the diffusion formulation, this target is identified with the denoised action distribution, i.e., $p_0(\cdot|s)=\pi_{\mathrm{MD}}(\cdot|s)$. Applying the forward diffusion process to $p_0$ then yields the corresponding noised marginal:
\begin{equation}
    p_t(a_t|s)
    =
    \int
    q_{t|0}(a_t|a_0)
    p_0(a_0|s)
    da_0.
\end{equation}
The DPMD reweighting function is
\begin{equation}
    g_{\mathrm{MD}}(a_t;s)
    =
    Z_{\mathrm{MD}}(s)\,p_t(a_t|s),
\end{equation}
which gives the marginal reweighted score-matching objective
\begin{equation}
    \mathcal{L}_{\mathrm{DPMD}}(\theta;s,t)
    =
    \int
    g_{\mathrm{MD}}(a_t;s)
    \left\|
        s_\theta(a_t;s,t)
        -
        \nabla_{a_t}\log p_t(a_t|s)
    \right\|^2
    da_t.
    \label{eq:app_dpmd_start}
\end{equation}
Applying the standard denoising score-matching equivalence yields
\begin{align}
    \mathcal{L}_{\mathrm{DPMD}}(\theta;s,t)
    &=
    Z_{\mathrm{MD}}(s)
    \int
    p_t(a_t|s)
    \left\|
        s_\theta(a_t;s,t)
        -
        \nabla_{a_t}\log p_t(a_t|s)
    \right\|^2
    da_t
    \nonumber\\
    &=
    Z_{\mathrm{MD}}(s)
    \iint
    q_{t|0}(a_t|a_0)
    \underbrace{
        \frac{
            \pi_{\mathrm{old}}(a_0|s)
            \exp\!\left(Q(s,a_0)/\lambda\right)
        }{
            Z_{\mathrm{MD}}(s)
        }
    }_{p_0(a_0|s)}
    \nonumber\\[-2mm]
    &\qquad\qquad
    \left\|
        s_\theta(a_t;s,t)
        -
        \nabla_{a_t}\log q_{t|0}(a_t|a_0)
    \right\|^2
    da_0da_t
    + C
    \nonumber\\
    &=
    \mathbb{E}_{
        a_0\sim\pi_{\mathrm{old}}(\cdot|s),\,
        a_t\sim q_{t|0}(\cdot|a_0)
    }
    \Bigg[
        \exp\!\left(Q(s,a_0)/\lambda\right)
    \nonumber\\[-1mm]
    &\qquad\qquad
        \left\|
            s_\theta(a_t;s,t)
            -
            \nabla_{a_t}\log q_{t|0}(a_t|a_0)
        \right\|^2
    \Bigg]
    + C.
    \label{eq:app_md_proof}
\end{align}
Thus, DPMD reduces policy mirror descent to the tractable denoising objective in Eq.~\ref{eq:app_md_proof}. Specifically, it samples $a_0\sim\pi_{\mathrm{old}}(\cdot|s)$ and $a_t\sim q_{t|0}(\cdot|a_0)$, and regresses toward the conditional score $\nabla_{a_t}\log q_{t|0}(a_t|a_0)$ with weight $\exp(Q(s,a_0)/\lambda)$.

\paragraph{Why same-state group normalization matters.}
States within a minibatch may differ substantially in difficulty and absolute value scale. Normalizing candidates across states would allow high-value actions from easy states to dominate the exponentiated weights, potentially suppressing the learning signal from difficult states. This is particularly problematic in hard states, where all candidates may have low absolute values but still exhibit meaningful relative differences. By normalizing within each state, GQRM estimates a local policy-improvement direction and converts these relative differences into effective supervision.

\textbf{GQRM derivation.} We define a group as a set of candidate actions independently sampled from $\pi_{\mathrm{old}}(\cdot|s)$ for a given state $s$. All candidates in the group share $s$ but differ in their sampled trajectory chunks. We evaluate them using $Q(s,a_0)$ and compute the normalization statistics within the group:
\begin{equation}
    \bar Q_G(s)
    =
    \mathbb{E}_{a_0\sim\pi_{\mathrm{old}}(\cdot|s)}
    \left[Q(s,a_0)\right],
    \qquad
    \sigma_G(s)
    =
    \sqrt{
    \mathbb{E}_{a_0\sim\pi_{\mathrm{old}}(\cdot|s)}
    \left[
        \big(Q(s,a_0)-\bar Q_G(s)\big)^2
    \right]
    }.
    \label{eq:app_group_stats_cont}
\end{equation}
The corresponding group-normalized value is
\begin{equation}
    \widetilde{Q}_G(s,a_0)
    =
    \mathrm{clip}\!\left(
        \frac{
            c\big(Q(s,a_0)-\bar Q_G(s)\big)
        }{
            \sigma_G(s)
        },
        -h,h
    \right),
    \label{eq:app_group_q}
\end{equation}
where $c$ controls the normalization scale and $h$ is the clipping threshold. The subscript $G$ emphasizes that both statistics are computed from candidates associated with the same state, rather than across unrelated states in a minibatch.

GQRM then redefines the KL-regularized policy-improvement target for each state $s$ by replacing the raw Q-score with the group-normalized score:
\begin{equation}
    \pi_{\mathrm{GQRM}}(\cdot|s)
    =
    \arg\max_{\pi}
    \;
    \mathbb{E}_{a_0\sim\pi(\cdot|s)}
    \big[\widetilde{Q}_G(s,a_0)\big]
    -
    \lambda
    D_{\mathrm{KL}}\!\left(
        \pi(\cdot|s)\,\|\,\pi_{\mathrm{old}}(\cdot|s)
    \right).
    \label{eq:app_gqrm_objective}
\end{equation}
Its closed-form solution is
\begin{equation}
    \begin{aligned}
    \pi_{\mathrm{GQRM}}(a_0|s)
    &=
    \frac{
        \pi_{\mathrm{old}}(a_0|s)
        \exp\!\left(\widetilde{Q}_G(s,a_0)/\lambda\right)
    }{
        Z_G(s)
    },
    \\
    Z_G(s)
    &=
    \int
    \pi_{\mathrm{old}}(a_0|s)
    \exp\!\left(\widetilde{Q}_G(s,a_0)/\lambda\right)
    da_0.
    \end{aligned}
    \label{eq:app_group_target}
\end{equation}
Identifying $p_{0,G}(\cdot|s)=\pi_{\mathrm{GQRM}}(\cdot|s)$ and applying the same forward diffusion process gives
\begin{equation}
    p_{t,G}(a_t|s)
    =
    \int
    q_{t|0}(a_t|a_0)
    p_{0,G}(a_0|s)
    da_0.
\end{equation}
Following the same derivation steps as DPMD, applying the same denoising score-matching equivalence yields the tractable GQRM objective:
\begin{align}
    \mathcal{L}_{\mathrm{GQRM}}(\theta;s,t)
    &=
    Z_G(s)
    \int
    p_{t,G}(a_t|s)
    \left\|
        s_\theta(a_t;s,t)
        -
        \nabla_{a_t}\log p_{t,G}(a_t|s)
    \right\|^2
    da_t
    \nonumber\\
    &=
    \mathbb{E}_{
        a_0\sim\pi_{\mathrm{old}}(\cdot|s),\,
        a_t\sim q_{t|0}(\cdot|a_0)
    }
    \Bigg[
        \exp\!\left(\widetilde{Q}_G(s,a_0)/\lambda\right)
    \nonumber\\[-1mm]
    &\qquad\qquad
        \left\|
            s_\theta(a_t;s,t)
            -
            \nabla_{a_t}\log q_{t|0}(a_t|a_0)
        \right\|^2
    \Bigg]
    + C.
    \label{eq:app_group_proof}
\end{align}
Thus, GQRM preserves the proposal distributions $\pi_{\mathrm{old}}(a_0|s),\,q_{t|0}(a_t|a_0)$ and the conditional score target $\nabla_{a_t}\log q_{t|0}(a_t|a_0)$, while replacing the raw DPMD regression weight $\exp(Q(s,a_0)/\lambda)$ with the group-normalized weight $\exp(\widetilde{Q}_G(s,a_0)/\lambda)$.

\paragraph{Monte Carlo implementation.}
For an $\epsilon$-prediction DDPM, conditional score matching is equivalent, up to a timestep-dependent scale, to predicting the injected Gaussian noise. In the finite-sample implementation, the statistics in Eq.~\ref{eq:app_group_stats_cont} are estimated from $N$ self-bootstrapped candidates sampled under the same observation. Each candidate is evaluated using the conservative clipped double-critic estimate $q_i=\min(Q_1,Q_2)(s,a_{0,i})$, giving
\begin{equation}
    \widetilde{Q}_G(s,a_{0,i})
    =
    \mathrm{clip}\!\left(
        \frac{
            c(q_i-\bar q_s)
        }{
            \sigma_s
        },
        -h,h
    \right),
    \quad
    \bar q_s=\frac{1}{N}\sum_{i=1}^{N}q_i,
    \quad
    \sigma_s=
    \sqrt{
        \frac{1}{N}
        \sum_{i=1}^{N}(q_i-\bar q_s)^2+\epsilon
    }.
    \label{eq:app_group_q_mc}
\end{equation}
Averaging Eq.~\ref{eq:app_group_proof} over states, same-state candidate groups, and forward diffusion noise yields the finite-sample GQRM objective:
\begin{equation}
    \mathcal{J}_{\mathrm{GQRM}}
    =
    \mathbb{E}_{s,\mathcal{G},\boldsymbol{\epsilon}}
    \left[
        \sum_{i=1}^{N}
        \exp\!\left(
            \widetilde{Q}_G(s,a_{0,i})/\lambda
        \right)
        \sum_{t=0}^{K_{\mathrm{ft}}-1}
        \left\|
            \boldsymbol{\epsilon}_{\phi'}
            (a_{t,i},t,s)
            -
            \boldsymbol{\epsilon}
        \right\|^2
    \right],
    \label{eq:app_gqrm_eps}
\end{equation}
where $\mathcal{G}=\{a_{0,i}\}_{i=1}^{N}$ is a group of candidates sampled for the same state, and $a_{t,i}$ is obtained by perturbing $a_{0,i}$ with noise $\boldsymbol{\epsilon}$ at diffusion timestep $t$. The actor is fine-tuned over the low-noise interval $t\in\{0,\ldots,K_{\mathrm{ft}}-1\}$. In practice, we retain the top-$k$ candidates ranked by their critic values and assign zero actor weight to candidates with $\widetilde{Q}_G(s,a_{0,i})\leq 0$. Both filtering operations, together with the group statistics and actor weights, are computed independently for each state. Although introduced for computational efficiency and variance reduction, this finite-sample implementation preserves the same underlying group-reweighting principle as the continuous formulation.


\section{Additional Training Details}
\label{app:implementation}

\begingroup
\setlength{\textfloatsep}{8pt}
\setlength{\floatsep}{8pt}
\setlength{\intextsep}{8pt}
\setlength{\abovecaptionskip}{4pt}
\setlength{\belowcaptionskip}{3pt}

\paragraph{Training scenes and parallel environments.}
We train on 56 GRScenes-100 scenes (47 home and 9 commercial), assigning one process to each scene. Each process runs eight synchronized IsaacLab environments, yielding 448 parallel environments. The processes are distributed as evenly as possible among Dingo, Unitree G1, and Unitree Go2. During training, each process keeps its assigned scene but periodically switches embodiments, retaining its local replay buffer and the DDP-synchronized actor and critics. The following sections describe the reward design, training schedule, rollout and replay construction, and MPC controller.

\subsection{Reward and Termination Design}
\label{app:reward}
The policy predicts a trajectory chunk at each decision step, and the MPC controller tracks this trajectory for a fixed 3-s interval before the next prediction. We treat each 3-s tracking interval as one RL transition. Its reward combines the simulator rewards accumulated during this interval with a path-progress reward computed at the end of the interval. The reward consists of six terms: a living regularizer, path progress, goal arrival, collision, immobility, and humanoid stability. The living regularizer applies a small penalty of $-0.025$ at each simulator step. The other terms are described below.

\textbf{Path progress.} We compute obstacle-aware progress using A$^\star$ search on the occupancy map. A free grid cell has traversal cost $c=\mathrm{clip}(1+0.1/(d_{\mathrm{obs}}+10^{-3}),1,1000)$, where $d_{\mathrm{obs}}$ is the cell's obstacle clearance in meters. This assigns a higher cost to cells near obstacles and therefore favors paths with greater clearance. Let $d_t$ be the minimum accumulated A$^\star$ path cost from the robot to the goal after tracking interval $t$, and define $\Delta_t=d_{t-1}-d_t$. Thus, $\Delta_t>0$ indicates progress. We set $h_t=1$ when the robot displacement has a positive dot product with its forward heading and $h_t=0$ otherwise. Combined with $\Delta_t$, this term rewards progress toward the goal only when the robot moves forward, rather than when it approaches the goal while moving backward. We use $\delta\psi_t$ for the absolute heading change over the interval and set $\eta=4.0$. Here, $\mathrm{clip}(x,a,b)=\min(\max(x,a),b)$. The progress reward is
\begin{equation}
\begin{aligned}
    r_{\mathrm{prog}}&=0.075\,\phi(\Delta_t,h_t,\delta\psi_t),\\
    \phi(\Delta_t,h_t,\delta\psi_t)&=
    \begin{cases}
        \mathrm{clip}(\Delta_t h_t,0,15), & \Delta_t>2,\\
        0, & \Delta_t<-1\ \land\ \delta\psi_t>\pi/3,\\
        0, & -1\leq\Delta_t\leq2\ \land\ \delta\psi_t>\pi/9,\\
        \mathrm{clip}\big(1.5(\Delta_t-\eta),-15,0\big), & \text{otherwise}.
    \end{cases}
\end{aligned}
    \label{eq:app_progress_reward}
\end{equation}
The first case rewards clear forward progress. The two zero-reward cases avoid penalizing large turns used for recovery. The final case penalizes insufficient progress.

\textbf{Goal arrival.} The arrival timer starts when the planar distance to the goal is below $0.5$\,m and the root linear speed is below $0.25$\,m/s. Once started, the timer runs for 4\,s before arrival is declared.

\textbf{Collision.} A collision penalty is applied when the contact-force magnitude exceeds $6.5$\,N or its increase exceeds $2.0$\,N. Collisions within 1\,m of the previous collision location are treated as repeated contacts, and their penalties are discounted by a factor of $0.9$ for each repetition.

\textbf{Immobility.} An immobility penalty is applied when the robot moves less than $0.2$\,m within a 6-s window.

\textbf{Humanoid stability.} For Unitree G1, a root height below $0.4$\,m latches the fall state and activates a stability penalty.

An episode terminates when any of the following occurs: the robot successfully arrives at the goal; the episode duration reaches 122\,s; the robot remains immobile for 12\,s; or, for Unitree G1, 4\,s elapse after a fall is latched. Collisions affect the reward but do not directly terminate the episode.

\begin{table}[H]
\centering
\caption{Reward components used for RL post-training. Each final reward term is the product of the listed raw signal and weight.}
\label{tab:app_reward_terms}
\scriptsize
\begin{tabular}{l l c}
\toprule
Term & Raw signal & Weight \\
\midrule
Living regularization & $1$ at each simulator step & $-0.025$ \\
Obstacle-aware progress & $\phi$ in Eq.~\ref{eq:app_progress_reward} & $0.075$ \\
Goal arrival & Distance- and velocity-conditioned shaping signal & $2.0$ \\
Collision & Non-positive, spatially discounted contact signal & $2.0$ \\
Immobility & $-1$ when displacement stays below $0.2$\,m over 6\,s & $0.2$ \\
Humanoid stability & $1$ after a Unitree G1 fall is latched & $-0.4$ \\
\bottomrule
\end{tabular}
\end{table}

\subsection{Training Schedule}
Training consists of three stages. \textbf{Stage 1: replay warmup.} During the first 3,000 vectorized simulator steps on each process, the environments collect replay transitions without gradient updates. \textbf{Stage 2: critic warmup.} After replay warmup, the first 100 minibatch optimizer updates train only the twin critics, while the actor remains fixed. \textbf{Stage 3: joint policy improvement.} Starting from optimizer update 101, the critics are updated on every minibatch and the actor is updated every two minibatch updates. Embodiment rotation continues throughout these stages. The first switch occurs at 30,000 simulator steps, followed by one switch every 24,000 steps; model parameters and local replay buffers are retained across switches.

After replay warmup, all processes synchronously perform one optimization round every 75 vectorized simulator steps. In each round, every process draws 192 transitions from its local replay buffer and splits them into eight minibatches of 24, resulting in eight optimizer updates. Gradients are synchronized across processes using distributed data parallelism. The actor and twin critics are optimized with Adam. Their learning rate is initialized to $2\times10^{-5}$ and linearly decayed to $1\times10^{-5}$ over 10,000 optimizer updates. We use $\gamma=0.99$ and $\tau=0.005$, update the target critics every two optimizer updates, and clip the gradient norm to $5.0$. For each actor update, we sample 64 candidates per state and score them using the minimum of the twin critics. Candidate values are centered within each state, normalized by the value dispersion across the minibatch, and clipped to $[-3,3]$. We retain the five highest-valued candidates with positive normalized scores and convert their scores into exponential weights using a learned temperature. The diffusion policy uses 10 denoising steps, of which only the six lowest-noise steps are fine-tuned.

\subsection{Chunk-Wise Rollout and Replay Construction}
During replay warmup, each environment executes a single perturbed candidate. After warmup, the policy generates eight unperturbed and eight perturbed candidates for each state. For evaluation episodes interleaved with training, the unperturbed candidate with the highest critic value is executed. For training episodes, the highest-valued perturbed candidate is selected with probability $0.5$; otherwise, one of the eight perturbed candidates is selected at random. Perturbed candidates are generated by the rollout exploration policy. Before execution, the selected candidate is truncated when necessary, corrected for kinematic feasibility, and smoothed.

After a trajectory chunk is executed, its macro-step reward is
\begin{equation}
    r_{\mathrm{macro}}
    =
    \frac{\sum_{\ell=0}^{L-1} r^{\mathrm{env}}_{t+\ell}+r_{\mathrm{prog}}}
         {L/(3\kappa_e)},
\end{equation}
where $L$ is the number of executed simulator steps and $\kappa_e\in\{1,2.5,2.5\}$ for Dingo, Unitree G1, and Unitree Go2, respectively. This normalization compensates for their different controller step frequencies and places full 3-s transitions on a comparable reward scale. Each replay transition stores the observation before planning, the executed trajectory, the macro-step reward, the terminal flag, and the embodiment identity. Each environment maintains a circular buffer of 250 transitions. The critic bootstraps from the next chunk-level observation in the same environment.

\subsection{MPC Tracking Controller}
MPC tracks each predicted trajectory and converts it into base velocity commands, which are executed by the embodiment-specific low-level controller. We model the robot base using the unicycle dynamics
\begin{equation}
    \dot{x}=v\cos\theta,\qquad
    \dot{y}=v\sin\theta,\qquad
    \dot{\theta}=\omega,
\end{equation}
with $|v|\leq0.5$\,m/s and $|\omega|\leq0.5$\,rad/s. Dingo uses 30 MPC steps at $0.1$\,s per step, while Unitree G1 and Unitree Go2 use 75 steps at $0.04$\,s per step; both settings provide a 3-s tracking horizon. The tracking objective uses state weights $\mathrm{diag}(10,10,0)$ and control weights $\mathrm{diag}(0.05,0.05)$, with the same state weights applied at the terminal step. The desired speed is $0.5$\,m/s and is reduced for short or highly curved trajectories, with a lower bound of $0.05$\,m/s. We solve the resulting nonlinear MPC problem using SQP with explicit Runge--Kutta integration.

\endgroup

\FloatBarrier
\section{Additional Experimental Results}
\label{app:additional-experiments}

\paragraph{Scene splits and ablation protocol.}
We conduct the main experiments on GRScenes-100 using a fixed 56/40 train-test split. The policy is RL post-trained on 56 scenes, comprising 47 home and 9 commercial scenes, and evaluated on 40 held-out scenes, comprising 20 home and 20 commercial scenes. Table~\ref{tab:main_sim} reports the results under this full training setting. All scenes are imported into IsaacLab with realistic collision geometry and physics simulation. Unless otherwise specified, simulation results are reported using Success Rate (SR) and Success weighted by Path Length (SPL) for three embodiments: Dingo as the wheeled robot, Unitree Go2 as the quadruped robot, and Unitree G1 as the humanoid robot. During RL training, rollout collection follows a synchronous execution protocol for efficiency: after the navigation policy predicts a waypoint trajectory, the low-level controller tracks it for a fixed 3-s interval before the next high-level policy inference. During evaluation, all policies are executed asynchronously: the navigation policy produces a new trajectory chunk whenever inference finishes, so the update interval is determined by the actual inference latency. To reduce the computational cost of component ablations, we train each variant on a 24-scene subset of the full training split, comprising 18 home and 6 commercial scenes. We evaluate these variants using the same metrics and embodiments as in the main experiments. These experiments are designed to isolate the contribution of individual components, while the final benchmark uses the full 56-scene training split.

\subsection{Cross-Embodiment Modulation Ablation}

We compare two ways of injecting robot embodiment information into the shared diffusion navigation policy. Soft-Prompt Concat directly feeds the learned robot embedding into the transformer decoder blocks as an additional conditioning token. In contrast, Embodiment FiLM Modulation maps the robot embedding to FiLM layers and uses them to modulate the output features of the transformer decoder blocks. We evaluate both strategies under the 24-scene ablation setting and the 56-scene full training setting to examine which conditioning strategy gains more from scaling the number of training scenes.

\begin{table}[H]
\centering
\caption{Cross-embodiment information injection ablation. Each cell reports SR/SPL. The 24-scene rows train on a subset of the 56 training scenes, while the 56-scene rows use the full RL training split.}
\label{tab:app_embodiment_modulation}
\scriptsize
\resizebox{\linewidth}{!}{
\begin{tabular}{l c c c c c c c}
\toprule
 & Overall & \multicolumn{2}{c}{Wheeled} & \multicolumn{2}{c}{Quadruped} & \multicolumn{2}{c}{Humanoid} \\
\cmidrule(lr){3-4}\cmidrule(lr){5-6}\cmidrule(lr){7-8}
Strategy & SR/SPL$\uparrow$ & Commercial & Home & Commercial & Home & Commercial & Home \\
\midrule
Soft-Prompt Concat (24 scenes) & 80.24/74.39 & 86.60/78.27 & 85.05/74.42 & 80.05/75.25 & 73.55/67.72 & 80.05/77.83 & 76.15/72.89 \\
Embodiment FiLM Modulation (24 scenes) & 80.48/74.78 & 85.80/79.48 & 85.75/76.49 & 78.45/73.66 & 74.25/68.69 & 81.85/78.13 & 76.75/72.25 \\
Soft-Prompt Concat (56 scenes) & 82.84/75.99 & 88.65/79.37 & 89.15/77.05 & 81.45/76.21 & 73.50/67.55 & 83.75/80.17 & 80.55/75.64 \\
Embodiment FiLM Modulation (56 scenes) & 84.28/77.19 & 88.55/78.96 & 88.70/76.37 & 80.65/75.72 & 79.05/72.35 & 84.25/80.41 & 84.50/79.37 \\
\bottomrule
\end{tabular}
}
\end{table}

Table~\ref{tab:app_embodiment_modulation} shows that both embodiment-conditioning strategies benefit from larger-scale co-training, but feature modulation scales better. Scaling Soft-Prompt Concat from 24 to 56 training scenes improves the overall score by 2.60 SR and 1.60 SPL, from 80.24/74.39 to 82.84/75.99. Scaling Embodiment FiLM Modulation improves the overall score by 3.80 SR and 2.41 SPL, from 80.48/74.78 to 84.28/77.19. Under the full 56-scene setting, Embodiment FiLM Modulation achieves the best overall performance and is especially helpful for the humanoid home split, where SR increases from 80.55 to 84.50 compared with Soft-Prompt Concat. This suggests that FiLM-style modulation of decoder features provides a stronger embodiment adaptation mechanism than directly passing the robot embedding into the transformer decoder blocks, especially when the shared policy is scaled to more scenes and must adapt clearance, turning, and recovery preferences across robot morphologies.

\subsection{Exploration Policy Ablation}

We ablate the Self-Bootstrapped Perturbation strategy used for structured exploration during RL post-training. As described in the main paper, this strategy samples a goal-conditioned trajectory and a goal-agnostic trajectory from the pretrained diffusion policy, mixes them with a signed coefficient, and applies coordinate-wise sign flips. The full exploration policy consists of three coupled components: goal-agnostic perturbation, signed trajectory reverse, and extrapolative mixing between goal-conditioned and goal-agnostic samples. We also evaluate a ``w/o extrapolation'' variant, where the two trajectories are combined with a fixed weight of 1. These ablations isolate whether the performance gain comes from structured self-generated candidate diversity rather than additional online data alone.

\begin{table}[H]
\centering
\caption{Exploration policy ablation in the 24-scene setting. Each cell reports SR/SPL.}
\label{tab:app_exploration_ablation}
\scriptsize
\resizebox{\linewidth}{!}{
\begin{tabular}{l c c c c c c c}
\toprule
 & Overall & \multicolumn{2}{c}{Wheeled} & \multicolumn{2}{c}{Quadruped} & \multicolumn{2}{c}{Humanoid} \\
\cmidrule(lr){3-4}\cmidrule(lr){5-6}\cmidrule(lr){7-8}
Variant & SR/SPL$\uparrow$ & Commercial & Home & Commercial & Home & Commercial & Home \\
\midrule
w/o reverse, only no-goal perturbation & 65.51/62.87 & 69.70/66.42 & 65.50/60.54 & 68.10/65.67 & 59.95/56.97 & 71.00/70.27 & 58.85/57.40 \\
w/o extrapolation & 77.84/71.74 & 84.60/77.30 & 85.90/76.07 & 75.90/70.52 & 73.25/67.16 & 75.40/71.77 & 72.00/67.67 \\
w/o no-goal perturbation and reverse & 6.23/4.75 & 4.86/3.87 & 1.65/1.30 & 9.15/8.86 & 5.40/5.09 & 9.10/5.43 & 7.25/4.00 \\
w/o no-goal perturbation, only reverse & 77.89/72.18 & 86.45/79.34 & 84.00/74.39 & 75.65/71.41 & 70.80/65.34 & 77.15/73.94 & 73.30/68.70 \\
Full exploration policy & 80.48/74.78 & 85.80/79.48 & 85.75/76.49 & 78.45/73.66 & 74.25/68.69 & 81.85/78.13 & 76.75/72.25 \\
\bottomrule
\end{tabular}
}
\end{table}

Table~\ref{tab:app_exploration_ablation} confirms that exploration quality is a primary bottleneck for RL post-training of pretrained diffusion navigation policies. Removing both goal-agnostic perturbation and trajectory reverse causes training to collapse, reducing overall SR/SPL to 6.23/4.75. Keeping either structured component recovers substantial performance: reverse alone reaches 77.89/72.18, and no-goal perturbation without reverse reaches 65.51/62.87. The full Self-Bootstrapped Perturbation policy performs best overall because it combines scene-consistent goal-agnostic samples with signed reverse and extrapolative mixing, producing lateral-shift, detour, and reverse-recovery candidates while preventing excessive deviation from the pretrained trajectory manifold. The ``w/o extrapolation'' variant also underperforms the full policy, indicating that fixed-weight mixing is less effective than extrapolative perturbation for reaching recovery behaviors outside the original goal-conditioned trajectory distribution.

\subsection{Closed-Loop RTC Guidance Ablation}

We further evaluate RTC guidance, which is enabled only during deployment to improve temporal consistency between consecutive trajectory predictions. This ablation uses the 56-scene training setting to measure the additional gain from this inference-time guidance on top of the full RL-trained policy.

\begin{table}[H]
\centering
\caption{RTC guidance ablation in the 56-scene setting. Each cell reports SR/SPL.}
\label{tab:app_rtc_ablation}
\scriptsize
\resizebox{\linewidth}{!}{
\begin{tabular}{l c c c c c c c}
\toprule
 & Overall & \multicolumn{2}{c}{Wheeled} & \multicolumn{2}{c}{Quadruped} & \multicolumn{2}{c}{Humanoid} \\
\cmidrule(lr){3-4}\cmidrule(lr){5-6}\cmidrule(lr){7-8}
Variant & SR/SPL$\uparrow$ & Commercial & Home & Commercial & Home & Commercial & Home \\
\midrule
w/o RTC & 82.17/75.81 & 86.40/78.12 & 87.60/75.90 & 77.70/73.33 & 75.45/69.42 & 83.80/80.55 & 82.10/77.54 \\
Full X-NavDP & 84.28/77.19 & 88.55/78.96 & 88.70/76.37 & 80.65/75.72 & 79.05/72.35 & 84.25/80.41 & 84.50/79.37 \\
\bottomrule
\end{tabular}
}
\end{table}

Table~\ref{tab:app_rtc_ablation} shows that RTC guidance provides a consistent but moderate improvement during closed-loop execution. Compared with the policy without RTC, the full method improves overall SR from 82.17 to 84.28 and SPL from 75.81 to 77.19. The gain is visible on most embodiment-scene splits, especially quadruped home scenes and humanoid home scenes, where temporal consistency helps reduce discontinuities between consecutive action chunks.


\end{document}